\def\year{2021}\relax
\documentclass[letterpaper]{article} 
\usepackage{aaai21}  
\usepackage{times}  
\usepackage{helvet} 
\usepackage{courier}  
\usepackage[hyphens]{url}  
\usepackage{graphicx} 
\usepackage{natbib}  
\usepackage{mathtools}
\usepackage{caption} 
\nocopyright
\usepackage{algorithm, algpseudocode}
\usepackage{algorithmicx}
\usepackage{xcolor}
\usepackage{calc}
\usepackage{multirow, makecell}
\usepackage[hang,flushmargin]{footmisc} 
\usepackage[switch]{lineno}

\definecolor{Crimson}{rgb}{0.86, 0.08, 0.24}
\definecolor{DarkGreen}{rgb}{0.00, 0.40, 0.00}
\definecolor{RoyalBlue}{rgb}{0.15, 0.25, 0.54}
\definecolor{DarkCyan}{rgb}{0.0, 0.54, 0.54}

\ifx \submission \undefined

\else

\fi

\newcommand{\figref}[1]{Figure~\ref{fig:#1}} 
\newcommand{\tabref}[1]{Table~\ref{tab:#1}}

\newcommand{\topic}[1]
{
\vspace{2mm} \noindent \textbf{#1}
}

\usepackage{bbm}
\usepackage{bm}

\usepackage{amsmath,amssymb}
\usepackage{booktabs}       
\usepackage{amsfonts}       
\usepackage{nicefrac}       
\usepackage{microtype}      
\usepackage{color}
\usepackage[labelformat=empty]{subfig}
\usepackage{array}
\usepackage{multirow}
\usepackage{amsthm}
\usepackage[para,online,flushleft]{threeparttable}

\newcommand{\comment}[1]{}

\newcommand{\cL}{{\mathcal{L}}}

\newcommand{\ie}{\textit{i}.\textit{e}.}

\newcommand{\mpage}[2]
{
\begin{minipage}{#1\linewidth}\centering
#2
\end{minipage}
}

\newcommand{\mfigure}[2]
{
\includegraphics[width=#1\linewidth]{#2}
}

\title{DropLoss for Long-Tail Instance Segmentation}

\author {
        Ting-I Hsieh\textsuperscript{\rm 1}\thanks{equal contribution},
        Esther Robb\textsuperscript{\rm 2}\footnotemark[1],
        Hwann-Tzong Chen\textsuperscript{\rm1,\rm3},
        Jia-Bin Huang\textsuperscript{\rm 2}\\
}

\affiliations {
    \textsuperscript{\rm 1} National Tsing Hua University \quad
    \textsuperscript{\rm 2} Virginia Tech \quad
    \textsuperscript{\rm 3} Aeolus Robotics 
}

\begin{document}

\maketitle
\vspace{-5mm}
\begin{abstract}

Long-tailed class distributions are prevalent among the practical applications of object detection and instance segmentation. 
Prior work in long-tail instance segmentation addresses the imbalance of losses between rare and frequent categories by reducing the penalty for a model incorrectly predicting a rare class label. 
We demonstrate that the rare categories are heavily suppressed by \textit{correct background predictions}, which reduces the probability for all foreground categories with equal weight. 
Due to the relative infrequency of rare categories, this leads to an imbalance that biases towards predicting more frequent categories. 
Based on this insight, we develop DropLoss -- a novel adaptive loss to compensate for this imbalance without a trade-off between rare and frequent categories. 
With this loss, we show state-of-the-art mAP across rare, common, and frequent categories on the LVIS dataset. Codes are available at \url{https://github.com/timy90022/DropLoss}.
\end{abstract}

\section{Introduction}
\label{introduction}

Object detection and instance segmentation have a wide array of practical applications.
State-of-the-art object detection methods adopt a multistage framework \cite{girshick2014rcnn,girshick2015fast,ren2015faster} trained on large-scale datasets with abundant examples for each object category \cite{lin2014coco}.
However, datasets used in real-word applications commonly fall into a long-tailed distribution over categories, i.e., the majority of classes have only a small number of training examples.
Training a model on these datasets inevitably induces an undesired bias towards frequent categories.
The limited diversity of rare-category samples further increases the risk of overfitting.
Methods for addressing the issues involving long-tailed distributions commonly fall into several groups: 
\emph{i}) resampling to balance the category frequencies, 
\emph{ii}) reweighting the losses of rare and frequent categories, and 
\emph{iii}) specialized architectures or feature transformations.

The instance segmentation problem presents unique challenges for learning long-tailed distributions, as it contains multiple training objectives to supervise region proposal, bounding box regression, mask regression, and object classification. 
Each of these losses contributes to the overall balance of model training.
The prior state-of-the-art in long-tail instance segmentation \citep{tan2020eql} discovered a phenomenon where the predictions for rare categories are suppressed by \emph{incorrect foreground class predictions}.
To reduce these ``discouraging gradients'' and allow the network to explore the solution space for rare categories, the EQL method~\citep{tan2020eql} removes losses to rare categories from incorrect foreground classification.
However, we observe that most ``discouraging gradients" in fact originate from \emph{correct background classification} (where a bounding box does not contain any labeled objects).
In the background case, the classification branch receives losses to suppress all foreground class prediction scores.

\begin{figure}[t]
\includegraphics[width=0.49\linewidth]{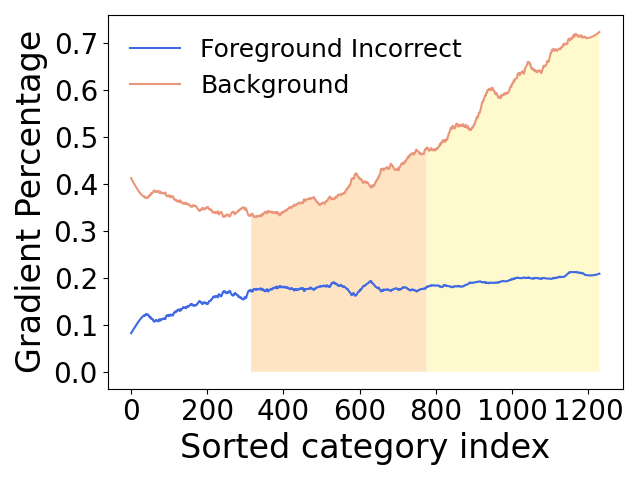} 
\includegraphics[width=0.49\linewidth]{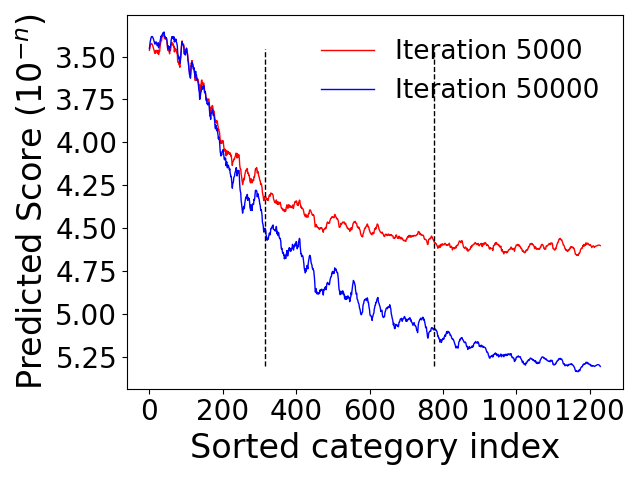}
\mpage{0.48}{(a) Gradient percentage}
\mpage{0.48}{(b) Prediction scores}
\caption{
Motivation.
  (a) Percentage of gradient updates from incorrect foreground classification (blue) and ground-truth background anchors (orange) on LVIS \cite{GuptaDG19}. 
  We divide the categories into ``frequent'' (white shading), ``common'' (orange shading), and ``rare'' (yellow shading). 
  For rare categories, background gradients occupy a disproportionate percentage of total gradients. 
  (b) The distribution of average foreground class prediction scores for ground-truth background bounding boxes at earlier (red) and later (blue) training stages. We find that, for background bounding boxes, the prediction scores of rare categories are more severely suppressed, and the training is biased towards predicting more frequent categories.
  } 
  \label{fig:motivation}
\end{figure}

In \figref{motivation}, we study the effect of such \emph{discouraging gradients} on the different categories of a long-tail dataset, categorized by number of training images into \emph{rare} (1-10 images), \emph{common} (11-100), and \emph{frequent} ($> 100$) categories.
We find that these losses disproportionately affect rare and common categories, due to the infrequency of ``encouraging gradients" in which a bounding box contains the correct category label.
Specifically, \figref{motivation}(a) shows that 50-70\% of discouraging gradients for rare categories originate from background predictions, compared with only 30-40\% of discouraging gradients for frequent categories.
Discouraging gradients from background classification (orange curve) contribute a much higher percentage of total discouraging gradients compared to that of incorrect foreground prediction (blue curve) as used in EQL \cite{tan2020eql}.
\figref{motivation}(b) shows that using a ground-truth background anchor, a trained model predicts scores for rare categories with several orders-of-magnitude lower confidence than for frequent categories.
This demonstrates a bias towards predicting more frequent categories.

Based on these observations, we develop a simple yet effective method to \emph{adaptively} rebalance the ratio of background prediction losses between rare/common and frequent categories.
Our proposed method \emph{DropLoss} removes losses for rare and common categories from background predictions based on sampling a Bernoulli variable with parameters determined by batch statistics.
DropLoss prevents suppression of rare and common categories, increasing opportunities for correct predictions of infrequent classes during training and reducing frequent class bias.

The contributions of this work are summarized as follows:
\begin{enumerate}
\item We provide an analysis of the unique characteristics of long-tailed distributions, particularly in the context of instance segmentation, to pinpoint the imbalance problem caused by disproportionate discouraging gradients from background predictions during training.
\item We develop a methodology for alleviating imbalances in the long-tailed setting by leveraging the ratio of rare and frequent classes in a sampled training batch.
\item We present state-of-the-art instance segmentation results on the challenging long-tail LVIS dataset \cite{GuptaDG19}.
\end{enumerate}

\section{Related Work}
\label{related_work}

\topic{Object Detection and Instance Segmentation.}
Two-stage detection architectures \cite{girshick2014rcnn,girshick2015fast,ren2015faster,lin2017pyramid} have been successful in the object detection setting, where the first stage proposes a ``region of interest'' and the second stage refines the bounding box and performs classification. 
This decomposition was initially proposed in R-CNN \cite{girshick2014rcnn}. 
Fast R-CNN \cite{girshick2015fast} and Faster R-CNN \cite{ren2015faster} improve efficiency and quality for object detection.
Mask R-CNN later adapts Faster R-CNN to the instance segmentation setting by adding a mask prediction branch in the second stage \cite{he2017mask}. 
Mask R-CNN has proven effective in a wide variety of instance segmentation tasks. Our work adopts this architecture. 
In contrast with two-stage methods, single-stage methods provide faster inference by eliminating the region proposal stage and instead predicting a bounding box directly from anchors \citep{liu2016ssd,redmon2016yolo,lin2017iccv}.
However, two-stage architectures generally provide better localization.

\topic{Learning Long-tailed Distributions.}
Techniques for learning long-tailed distributions generally fall into three groups: 
resampling, 
reweighting and cost-sensitive learning, 
and feature manipulation. 
We discuss each in the following sections.

\topic{Resampling Methods.} 
Oversampling methods \citep{chawla2002smote,han2005borderline,mahajan2018exploring,masko2015impact,huang2016learning,he2008adasyn,zou2018unsupervised} duplicate rare class samples to balance out the class frequency distribution. 
However, oversampling methods tend to overfit to the rare categories, as this type of method does not address the fundamental lack of data. 
Several oversampling methods aim to address this by augmenting the available data \cite{chawla2002smote,han2005borderline}, but undersampling methods are often preferred \cite{drummond2003c4}.
Undersampling methods \citep{drummond2003c4,tsai2019under,kahn1953methods} remove frequent class samples from the dataset to balance the class frequency distribution. 
The loss of information from removing these samples can be mitigated through careful selection using statistical techniques \citep{tsai2019under,kahn1953methods}.
It can be beneficial to combine the advantages of undersampling and oversampling \cite{chawla2002smote}.
Dynamic methods adjust the sampling distribution throughout training based on loss or metrics \citep{pouyanfar2018dynamic}.
Class balance sampling \citep{kang2019decoupling,shen2016relay} uses class-aware strategies to rebalance the data distribution for learning classifiers and representations.
In the context of the dense instance segmentation problem, it is difficult to apply the above resampling methods because the number of class examples per image may vary.

\topic{Reweighting and Cost-sensitive Methods.}
Rather than rebalancing the sampling distribution, reweighting methods seek to balance the \emph{loss weighting} between rare and frequent categories.
Class frequency reweighing methods commonly use the inverse frequency of each class to weight the loss \citep{huang2016learning,wang2017learning,cui2019class}.
Cost-sensitive methods \citep{li2019gradient,lin2017focal} aim to balance the model loss magnitudes between rare and frequent categories. 
An existing meta-learning method \cite{shu2019meta} explicitly learns loss weights based on the data. 
Our method provides a simple way to combine class frequency-aware sampling and cost-sensitive learning.

\topic{Feature Manipulation Methods.}
In contrast to resampling methods and reweighting methods that focus on modifying the loss based on class frequency, feature manipulation methods aim to design specific architectures or feature relationships to address the long-tail problem. 
Normalization can be used to control the distribution of deep features, preventing frequent categories from dominating training \cite{kang2019decoupling}. 
Metric learning methods \cite{kang2019decoupling,zhang2017range} learn maximally-distant ``prototypes'' of deep features to improve performance on data-scarce categories, effectively transferring knowledge between head and tail categories. 
Similarly, knowledge transfer in feature space can be accomplished using memory-based prototypes \cite{liu2019large} or transfer of intra-class variance \citep{yin2019feature}.

\topic{Long-tail Learning Settings.}
Several methods have been proposed to handle the problem of learning from imbalanced datasets in other settings such as object classification~\cite{cui2019class,cao2019learning,jamal2020rethinking,tan2020eql}.
In the long-tail object recognition setting, the prior state-of-the-art method \cite{tan2020eql} uses selective reweighting. 
Their work observed that rare categories receive significantly more ``discouraging gradients'' compared with frequent categories, and develop a method for rebalancing discouraging gradients from foreground misclassifications.
Their method uses a binary 0 or 1 reweighting based on whether the class is rare or frequent. 
Unlike this work, our method focuses on the much more prevalent background classification losses in the instance segmentation setting, and we develop a new adaptive resampling and reweighting method which accounts for this imbalance and for the distribution of classes within a sample.
Note that most of the methods for learning long-tailed distributions focus on the \emph{image classification} setting where this is no background class (and therefore no losses associated with background). 
In object detection and instance segmentation, however, the background class plays a very dominant role in the loss.
This inspires our design of a reweighting mechanism which specifically considers background class.

\section{Method}
\label{headings}


Based on the observation that rare and common categories receive disproportionate discouraging gradients from background classifications (compared with frequent categories), the goal of our method is to prevent rare categories from being overly suppressed by reducing the imbalance of discouraging gradients. 
We are inspired by work in one-stage object detection \cite{li2019gradient,lin2017focal}, which encounters a similar problem of large gradients from negative anchors inhibiting learning.
We first construct a baseline which modifies the equalization loss \cite{tan2020eql} to rebalance gradients from foreground and \emph{background} region proposals for rare/common and frequent categories.
We show that this baseline leads to improved results over \cite{tan2020eql} but requires careful hyperparameter selection and exhibits a clear tradeoff between rare/common and frequent categories.
To alleviate this problem, we propose a stochastic  \emph{DropLoss} which improves the overall frequent-rare category performance as well as improving the tradeoff as measured by a Pareto frontier \ref{fig:parteo}.

\begin{figure}[t]
\includegraphics[clip,width=0.49\linewidth]{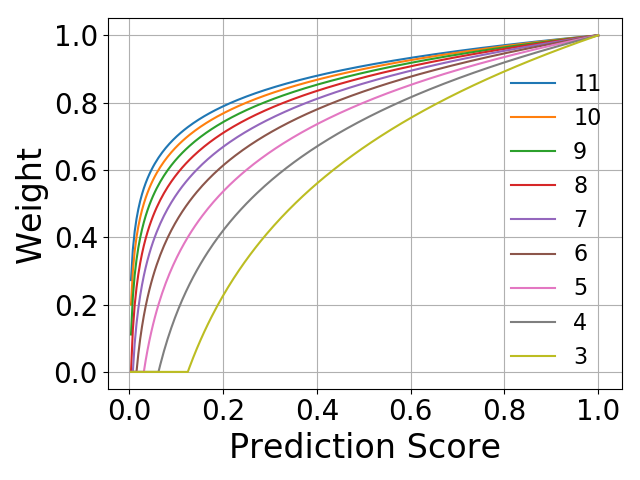}  
\includegraphics[clip,width=0.49\linewidth]{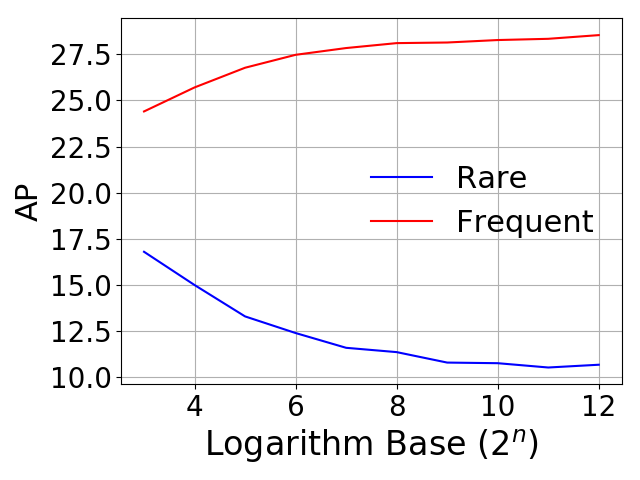} 
\mpage{0.45}{(a) Weight $w_j$ as a function of the logarithm base $b$.}
\mpage{0.45}{(b) Frequent-rare category performance tradeoff}
  \caption{Background equalization loss. We present an extension of the equalization loss that specifically focuses on the background classification.
  (a) The curves of the $\cL_\mathrm{BEQL}$ weights in \eqref{eqn:BEQL_weight} with different choices for the logarithm base $b$. Smaller values of the logarithm base $b$ reduce the effects of background more.
  (b) The experimental result of applying different logarithm bases shows a tradeoff between the mean average precision (mAP) of rare categories and the mAP of frequent categories with respect to different logarithm base settings. 
  Note that the background equalization loss $\cL_\mathrm{BEQL}$ with a large base $b$ reduces to the existing equalization loss $\cL_\mathrm{EQL}$.
  }
  \label{fig:BEQL}
\end{figure}

\subsection{Revisiting the Equalization Loss}
\topic{Equalization Loss.}
We start with a review of the equalization loss \cite{tan2020eql}. 
Equalization loss modifies sigmoid cross-entropy to alleviate discouraging gradients from incorrect foreground predictions. 
Note that, in sigmoid cross-entropy, the ground-truth label $y_j$ represents only a binary distribution for the foreground category $j$, and no extra class label for the background is included. 
That is, we have $y_j=1$ if the ground-truth category of a region is $j$. 
On the other hand, if a region belongs to the background, we have $y_j=0$ for all the categories. 
During training, a region proposal is labeled as ``background" if its IoU with any ground-truth region of a foreground class is lower than 50\%.

Given a region proposal $r$, the equalization loss is formulated as follows: 
\begin{equation}
    \cL_\mathrm{EQL} = -\sum_{j=1}^C w_j \log(\hat{p}_j) \,,
    \label{eqn:EQL}
    \hspace{0.15cm}
    \hat{p}_j= 
    \begin{cases}
    p_j,& \text{if~} y_j=1 \,, \\
    1-p_j, & \text{otherwise,}
    \end{cases}
\end{equation}
\begin{equation}
    \label{eqn:EQL_weight}
    w_j = 1 - E(r)T_{\lambda}(f_j)(1-y_j) \,, 
\end{equation}
%
where $C$ is the number of categories, $p_j$ is predicted logit, and $f_j$ is the frequency of category $j$ in the dataset. 
The indicator function $E(r)$ outputs $1$ if $r$ is a foreground region and $0$ if it belongs to the background. 
More specifically, for a region proposal $r$ that is considered a background region with all $y_j$ being zero, we have $E(r)=0$.
$T_\lambda(f)$ is also a binary indicator function that, given a threshold $\lambda$, outputs $1$ if $f<\lambda$ to indicate the category is of low frequency. 
It can be verified from \eqref{eqn:EQL_weight} that, for a foreground region $r$ (\ie, $E(r)=1$), the weight $w_j$ is either $1$ or $1-T_\lambda(f_j)$, depending on the frequency of the ground-truth category $j$. 
Further, if a category $j$ is of low frequency (rare category), the weight $w_j=1-T_\lambda(f_j)$ becomes zero and thus no penalty is given to incorrect foreground predictions.
On the other hand, for frequent categories, the weight is $1$ and the penalty of incorrect prediction remains $-\log(1-p_j)$. 
By removing discouraging gradients to rare/common categories from incorrect foreground predictions, the equalization loss achieved state-of-the-art on the LVIS Challenge 2019.
Foreground class label prediction is selected using the maximum logit, so entirely removing the loss allows the network to optimize rare categories without penalties, as long as the prediction logit is less than ground truth for the frequent categories. 
This approach removes large penalties for non-zero confidences in rare categories, which otherwise imbalance the training to suppress rare categories.

%

\topic{Background Equalization Loss.} 
In contrast to the mechanism of equalization loss that prevents large penalties for non-zero confidences in rare categories, cost-sensitive learning methods only reduce \cite{lin2017focal} or remove \cite{li2019gradient} discouraging gradients if the magnitude of the loss falls below some threshold.  
Our core insight is that foreground and background categories require different approaches due to the differences in prediction criteria. 
For background categories, the network predicts background class if \emph{all} logits $p_j$ fall below a threshold. 
For foreground categories, the prediction is selected using the \emph{maximum} logit $p_j$. 
Inspire by cost-sensitive loss and equalization loss, we present the \emph{background equalization loss} as an extension to the original equalization loss:
\begin{equation}
    \cL_\mathrm{BEQL} = -\sum_{j=1}^C w_j \log(\hat{p}_j) \,,
    \label{eqn:BEQL}
    \hspace{0.15cm}
    \hat{p}_j= 
    \begin{cases}
    p_j,& \text{if~} y_j=1 \,, \\
    1-p_j, & \text{otherwise,}
    \end{cases}
\end{equation}
\begin{equation}
    \label{eqn:BEQL_weight}
    w_j= 
    \begin{cases}
    1 - T_{\lambda}(f_j)(1-y_j),& \text{if } E(r) = 1 \,, \\
    1 - T_{\lambda}(f_j) \cdot \min\{-\log_b(p_j), 1\}, & \text{otherwise.}
    \end{cases}
\end{equation}
\noindent
By comparing \eqref{eqn:EQL_weight} and \eqref{eqn:BEQL_weight},
we can see that the background equalization loss differs from the equalization loss in the weights for background regions. 
The equalization loss always penalizes a background region ($E(r)=0$ and thus $w_j=1$) even if the category is of low frequency. 
In contrast, our background equalization loss gives smaller weight to background predictions as long as their confidences are low.
We use a logarithm base $b$ to control the sensitivity of the weight concerning the confidence of background prediction. 
\figref{BEQL}(a) shows the curves of the $\cL_\mathrm{BEQL}$ weights in \eqref{eqn:BEQL_weight} by varying the value of the logarithm base $b$. 
For example, suppose we would want to focus on the performance of the rare category, we can set the value of $b=2$.
The main idea here is to alleviate the accumulation of small but non-negligible discouraging gradients from the background. 
When applying the proposed background equalization loss with different logarithm bases, however, we see a clear performance tradeoff between frequent and rare categories (see \figref{BEQL}(b)). 
The results show that the average precision of the rare categories behaves in the \emph{opposite} way as the average precision of the frequent categories for different choices of logarithm bases. 

%

\subsection{DropLoss}
While suppressing discouraging gradients from the background shows improvement for the rare categories, the background equalization loss has a drawback.
The performance often sensitively depends on the choice of the logarithm base. 
It is difficult to choose an appropriate logarithm base that works for different long-tailed distributions without suffering from a tradeoff between frequent and rare categories. 
In light of this, we propose a new stochastic method, called \emph{DropLoss}, which dynamically balances the influence of background discouraging gradients for rare/common/frequent categories.

\begin{table*}[t]
\centering
\resizebox{\textwidth}{!}{%
\begin{tabular}{lllcccccccc}
\toprule
Architecture & Backbone & Loss & AP (\%) & AP$_\mathrm{50}$ & AP$_\mathrm{75}$ & AP$_\mathrm{r}$ & AP$_\mathrm{c}$ & AP$_\mathrm{f}$ & AR & AP$_\mathrm{bbox}$ \\ \midrule
\multirow{3}{*}{Mask R-CNN} & \multirow{3}{*}{R-50-FPN} & BCE & 21.5 & 33.4 & 22.9 & 4.7 & 21.2 & \textbf{28.6} & 28.3 & 21 \\
 &  & EQL~\cite{tan2020eql} & 23.8 & 36.3 & 25.2 & 8.5 & 25.2 & 28.3 & 31.5 & 23.5 \\
 &  & DropLoss (Ours) & \textbf{25.5} & \textbf{38.7} & \textbf{27.2} & \textbf{13.2} & \textbf{27.9} & 27.3 & \textbf{34.8} & \textbf{25.1} \\ \midrule
\multirow{3}{*}{Mask R-CNN} & \multirow{3}{*}{R-101-FPN} & BCE & 23.6 & 36.5 & 25.1 & 5.6 & 24.2 & \textbf{30.1} & 30.9 & 23.3 \\
 &  & EQL~\cite{tan2020eql} & 26.2 & 39.5 & 27.9 & 11.9 & 27.8 & 29.8 & 33.8 & 26.2 \\
 &  & DropLoss (Ours) & \textbf{26.9} & \textbf{40.6} & \textbf{28.9} & \textbf{14.8} & \textbf{29.7} & 28.3 & \textbf{36.4} & \textbf{26.8} \\ \midrule
\multirow{3}{*}{Cascade R-CNN} & \multirow{3}{*}{R-50-FPN} & BCE & 21.4 & 32 & 23.1 & 3.4 & 20.4 & \textbf{29.8} & 27.6 & 22.8 \\
 &  & EQL~\cite{tan2020eql} & 24.2 & 35.9 & 25.8 & 7.8 & 25 & 29.7 & 31.4 & 26 \\
 &  & DropLoss (Ours) & \textbf{25} & \textbf{37} & \textbf{26.9} & \textbf{9.1} & \textbf{27.2} & 28.7 & \textbf{34} & \textbf{26.9} \\ 
 \midrule
\multirow{3}{*}{Cascade R-CNN} & \multirow{3}{*}{R-101-FPN} & BCE & 23 & 34.4 & 24.7 & 3.5 & 22.8 & \textbf{31.2} & 29.9 & 24.9 \\
 &  & EQL~\cite{tan2020eql} & 25.4 & 37.3 & 27.3 & 7.2 & 26.6 & 31 & 33.1 & 27.2 \\
 &  & DropLoss (Ours) & \textbf{26.4} & \textbf{39} & \textbf{28.1} & \textbf{11.5} & \textbf{28.5} & 29.7 & \textbf{35.5} & \textbf{28.6} \\ \bottomrule
\end{tabular}%
}
\caption{Comparison between architecture and backbone settings, evaluated on LVIS v0.5 validation set. We compare BCE (binary cross-entropy), EQL (equalization loss) and Drop (DropLoss). AP/AR refers to mask AP/AR, and subscripts `r’, `c’, and `f’ refer to rare, common, and frequent categories. }
\label{tab:LVIS Effectiveness}
\end{table*}

Similar to the design of the background equalization loss, we seek to adjust weights on the logits of low-frequency categories for background region proposals. 
In DropLoss, we introduce a Bernoulli distribution and sample a binary value from the distribution as the weight $w_j$ if a region belongs to the background. 
Further, we determine the parameter of the Bernoulli distribution by a beta sampling distribution over the occurrence ratios of rare, common, and frequent categories for the regions generated by the Region Proposal Network \cite{ren2015faster} during training. 
The Bernoulli distribution with a Beta prior is suitable because we aim to model binary outcomes with varying biases in a stochastic manner.

Given a batch of region proposals, we compute the ratio between the occurrences of `rare + common' categories to all foreground occurrences (\ie, `rare + common + frequent' categories). 
In other words, we treat a batch of region proposals as a sample of occurrence ratio that is drawn from a beta distribution to provide the parameter of the Bernoulli distribution. 
Our intuition behind such a scheme is simple: 
For region proposals of rare and common categories, their occurrences in a batch are of low frequency.
Therefore, the discouraging gradients from the background predictions should be accordingly discounted for rare and common categories. 

We formulate DropLoss as follows:
\begin{equation}
    \cL_\mathrm{Drop} = -\sum_{j=1}^C w_j \log(\hat{p}_j) \,,
    \label{eqn:Drop}
    \hspace{0.15cm}
    \hat{p}_j= 
    \begin{cases}
    p_j,& \text{if } y_j=1 \,, \\
    1-p_j, & \text{otherwise,}
    \end{cases}
\end{equation}
\begin{equation}
    w_j= 
    \begin{cases}
    1 - T_{\lambda}(f_j)(1-y_j),& \text{if } E(r) = 1 \,,\\
    w \sim \mathrm{Ber}(\mu_{f_j}), & \text{otherwise,}
    \end{cases}
\end{equation}
\noindent 
where a random sample $w \in \{0,1\}$ is drawn from Bernoulli distribution $\mathrm{Ber}(\mu_{f_j})$ if the region proposal $r$ belongs to the background, \ie, $E(r)=0$. 
The parameter $\mu_{f_j}$ of the Bernoulli distribution is determined by the occurrence ratio of low-frequency (`rare + common') categories in the current batch of region proposals. 
We compute the parameter by
\begin{equation}
    \mu_{f_j}=
    \begin{cases}
    (n_\mathrm{rare} + n_\mathrm{common})/n_\mathrm{all},& \text{if } T_{\lambda}(f_j)=1 \,,\\
    n_\mathrm{frequent} /n_\mathrm{all}, & \text{otherwise,}
    \end{cases}
\end{equation}
\noindent
where $n_\mathrm{rare}$, $n_\mathrm{common}$, and $n_\mathrm{frequent}$ are the numbers of occurrences of rare, common, and frequent categories in the current training batch of foreground region proposals. 
The total number of foreground occurrences is $n_\mathrm{all}=n_\mathrm{rare}+n_\mathrm{common}+n_\mathrm{frequent}$.
Implementation of the above DropLoss scheme is straightforward: 
For each batch, we derive the parameter $\mu_{f_j}$ depending on whether category $j$ is rare/common or frequent. 
We can then simulate a flip of a biased coin with head probability $\mu_{f_j}$ and assign $w_j=1$ if we get a head. 

A region proposal is annotated as a background region if it does not overlap with any ground-truth foreground region, or if the IoU is lower than 50\%. If the number of rare category occurrences in a given batch is large, discouraging gradients to that rare category are more likely to be kept (with a higher chance to get $w_j=1$).
On the other hand, if a rare category does not appear very often in a batch, it is highly probable that discouraging gradients to the rare category will be dropped. 
Therefore, our dropping strategy tends to neglect unrelated non-overlapping background proposals but would be inclined to keep more related ($0<$ IoU $< 0.5$) background proposals.


\section{Experimental Results}
\label{others}

In this section, we present the implementation details and experimental results. 
We compare DropLoss with the state-of-the-art long-tail instance segmentation baselines on the challenging LVIS dataset \cite{GuptaDG19}. 
To validate the effectiveness of this approach, we compare across different architectures and backbones and integrate with additional long-tail resampling methods. 
We find that DropLoss demonstrates consistently improved results in AP and AR across all these experimental settings.

\topic{Dataset.}
Following the previous work \emph{equalization loss} \citep{tan2020eql}, we train and evaluate our model on LVIS benchmark dataset. 
LVIS is a large vocabulary instance segmentation dataset, containing 1{,}230 categories.
In LVIS dataset, categories are sorted into three groups based on the number of images
in which they appear: \textit{rare} (1-10 images), \textit{common} (11-100), and \textit{frequent} ($>100$). 
We report AP for each bin to quantify performance in the long-tailed distribution setting. 
We train our model on the 57K-image LVIS v0.5 training set and evaluate it on the 5K-image LVIS v0.5 validation set.

\begin{figure}[t!]
\includegraphics[clip,width=0.49\linewidth]{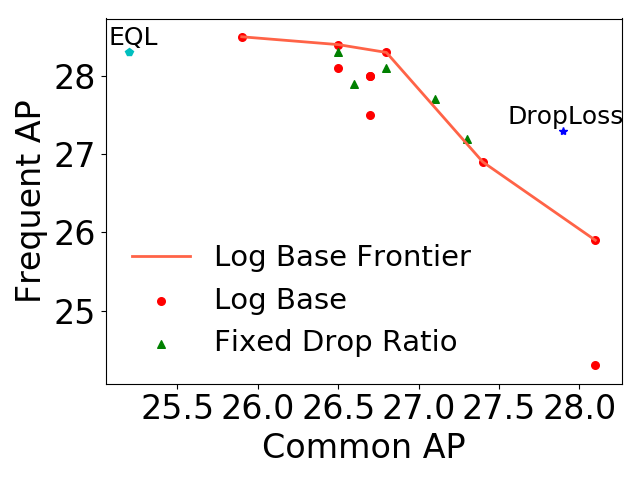} \hfill
\includegraphics[clip,width=0.49\linewidth]{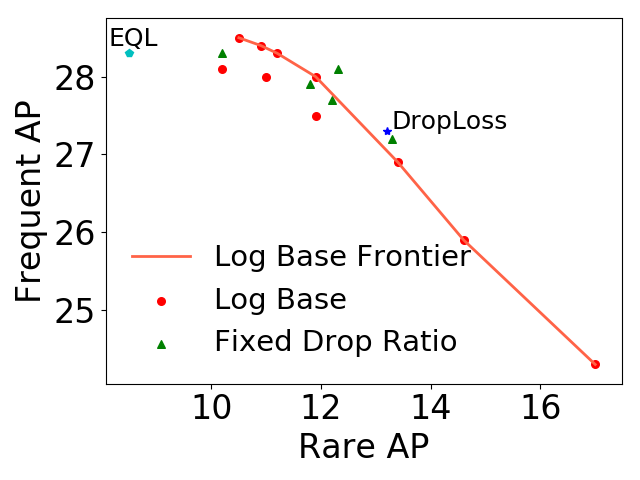}
\caption{Measuring the performance tradeoff. 
  Comparison between rare, common, and frequent categories AP for baselines and our method. We visualize the tradeoff for `common vs. frequent' and `rare vs. frequent'as a Pareto frontier, where the top-right position indicates an ideal tradeoff between objectives. DropLoss achieves an improved tradeoff between object categories, resulting in higher overall AP.}
  \label{fig:parteo}
\end{figure}

\topic{Implementation Details.}
For our experiments, we adopt the Mask R-CNN \citep{HeGDG20} architecture with Feature Pyramid Networks \citep{lin2017feature} as a baseline model. 
We train the network using stochastic gradient descent with a momentum of $0.9$ and a weight decay of $0.0001$ for 90K iterations, with batch size 16 on eight parallel NVIDIA 2080 Ti GPUs. 
We initialize the learning rate to $0.2$ and decay it by a ratio of $0.1$ at iterations 60{,}000 and 80{,}000. 
We use the Detectron2 \cite{wu2019detectron2} framework with default data augmentation. 
The data augmentation includes scale jitter with a short edge of  (640, 672, 704, 736, 768, 800) pixels and a long edge no more than 1{,}333 pixels horizontal flipping. 
In the Region Proposal Network (RPN), we sample 256 anchors with a 1:1 ratio between foreground and background to compute the RPN loss and choose 512 ROI-aligned proposals per image with a 1:3 foreground-background ratio for later predictions. 
Based on LVIS \citep{GuptaDG19}, the prediction threshold is reduced from $0.05$ to $0.0$, and we set the top 300 bounding boxes as prediction results. This setting is widely used in LVIS training and evaluation.
For all the experiments, we report the average results of three independent runs of model training. 
The variances in AP are generally small (approximately 0.1-0.2).

\begin{table*}[t!]
\centering
\resizebox{\textwidth}{!}{%
\begin{tabular}{lccccccccccc}
\toprule
 Method & Use RFS & AP (\%) & AP$_\mathrm{50}$ & AP$_\mathrm{75}$ & AP$_\mathrm{r}$ & AP$_\mathrm{c}$ & AP$_\mathrm{f}$ & AP$_\mathrm{s}$ & AP$_\mathrm{m}$ & AP$_\mathrm{L}$ & AP$_\mathrm{bbox}$ \\ \midrule
Sigmoid & - & 21.5 & 33.4 & 22.9 & 4.7 & 21.2 & \textbf{28.6} & 15.6 & 29.3 & 39 & 21 \\
Softmax & - & 21.3 & 33.1 & 22.6 & 3 & 21.2 & \textbf{28.6} & 15.8 & 28.5 & 39.2 & 21 \\
EQL~\cite{tan2020eql} & - & 23.8 & 36.3 & 25.2 & 8.5 & 25.2 & 28.3 & 17.1 & 31.4 & 41.7 & 23.5 \\
DropLoss (Ours) & -  & \textbf{25.5} & \textbf{38.7} & \textbf{27.2} & \textbf{13.2} & \textbf{27.9} & 27.3 & \textbf{17.7} & \textbf{32.7} & \textbf{43.2} & \textbf{25.1} \\ \midrule
Sigmoid & \checkmark & 23.8 & 36.3 & 25.2 & 8.5 & 25.2 & \textbf{28.3} & 17.1 & 31.4 & 41.7 & 23.5 \\
Softmax & \checkmark & 24.3 & 37.8 & 25.9 & 14.1 & 24.3 & \textbf{28.3} & 16.5 & 31.6 & 41.2 & 23.8 \\
EQL~\cite{tan2020eql} & \checkmark & 25.5 & 39 & 27.2 & 16.7 & 26.3 & 28.1 & 17.5 & 33 & 43 & 25 \\
DropLoss (Ours) & \checkmark & \textbf{26.4} & \textbf{40.3} & \textbf{28.4} & \textbf{17.3} & \textbf{28.7} & 27.2 & \textbf{17.9} & \textbf{33.1} & \textbf{44} & \textbf{25.8} \\ 
\bottomrule
\end{tabular}%
}

\resizebox{\textwidth}{!}{%
\begin{tabular}{lccccccccccc}
\toprule
 Method & Use RFS & AP (\%) & AP$_\mathrm{50}$ & AP$_\mathrm{75}$ & AP$_\mathrm{r}$ & AP$_\mathrm{c}$ & AP$_\mathrm{f}$ & AP$_\mathrm{s}$ & AP$_\mathrm{m}$ & AP$_\mathrm{L}$ & AP$_\mathrm{bbox}$ \\ \midrule
Baseline & - & 16.2 & 25.9 & 16.9 & 0.7 & 12.6 & 27 & 10.5 & 22.7 & 32.7 & 16.6 \\
EQL~\cite{tan2020eql} & - & 18.4 & 28.6 & 19.4 & 2.5 & 16.5 & \textbf{27.4} & 11.9 & 25.4 & 35.6 & 18.9 \\
DropLoss (Ours) & -  & \textbf{19.8} & \textbf{30.9} & \textbf{20.9} & \textbf{3.5} & \textbf{20} & 26.7 & \textbf{12.9} & \textbf{27.5} & \textbf{37.1} & \textbf{20.4} \\ \midrule
Baseline & \checkmark & 18.8 & 29.6 & 19.9 & 5.6 & 16.6 & 27.1 & 11.6 & 25.6 & 35.7 & 19.2 \\
EQL~\cite{tan2020eql} & \checkmark & 21 & 32.7 & 22.3 & 9.1 & 20.1 & \textbf{27.3} & 13.1 & 28.5 & 39.2 & 21.7 \\
DropLoss (Ours) & \checkmark & \textbf{22.3} & \textbf{34.5} & \textbf{23.6} & \textbf{12.4} & \textbf{22.3} & 26.5 & \textbf{13.9} & \textbf{29.9} & \textbf{40} & \textbf{22.9} \\ 
\bottomrule
\end{tabular}%
}
\caption{Evaluation on LVIS v0.5 (top) and LVIS v1.0 (bottom) validation sets \emph{with} and \emph{without} Repeat Factor Sampling (RFS). Here we use Mask-RCNN and ResNet-50. DropLoss achieves the best overall AP across both settings. 
}
\label{tab:compare}
\end{table*}

\begin{figure*}[t]
    \mfigure{0.24}{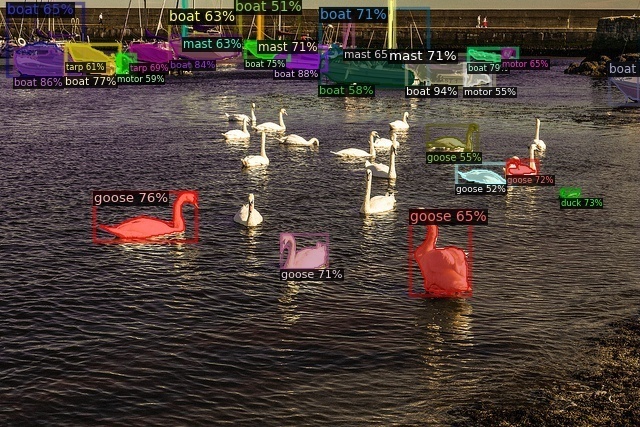} \hfill
\includegraphics[height=0.16\linewidth, width=0.24\linewidth]{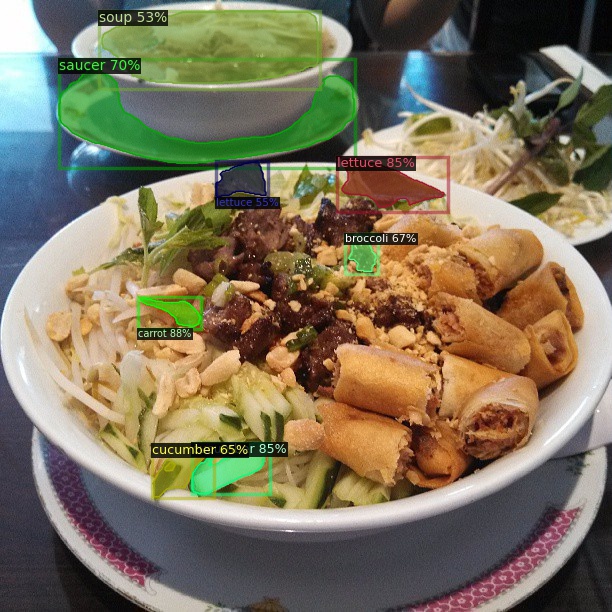} \hfill
    \mfigure{0.24}{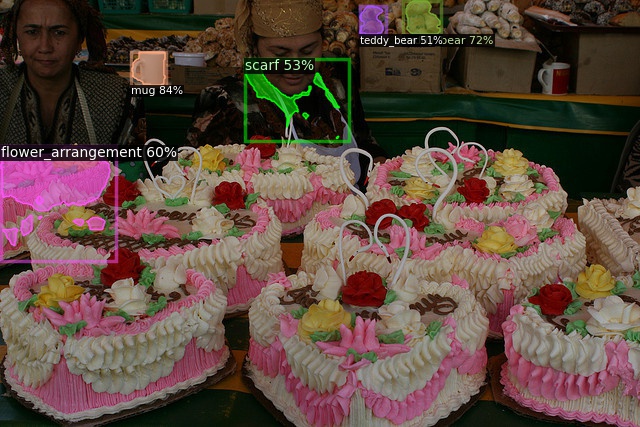} \hfill
    \mfigure{0.24}{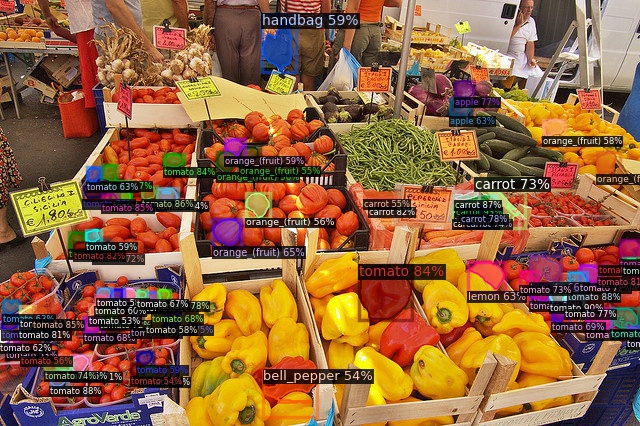}\vspace{-2mm}\\
$\underbracket[1pt][2.0mm]{\hspace{\textwidth}}_%
    {\substack{\vspace{-3.0mm}\\\colorbox{white}{~~Mask RCNN with softmax loss~~}}}$\vspace{1mm}\\%

    \mfigure{0.24}{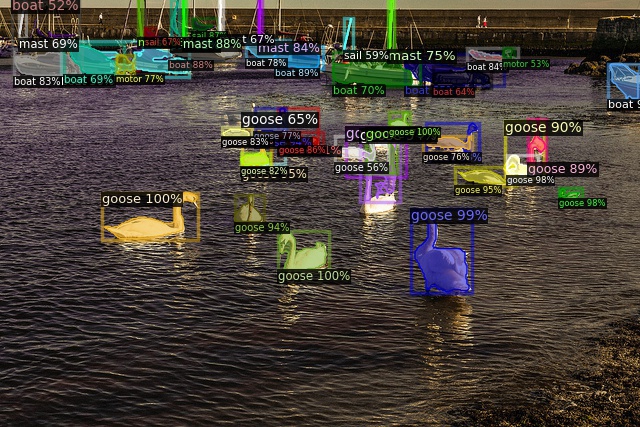} \hfill
\includegraphics[height=0.16\linewidth, width=0.24\linewidth]{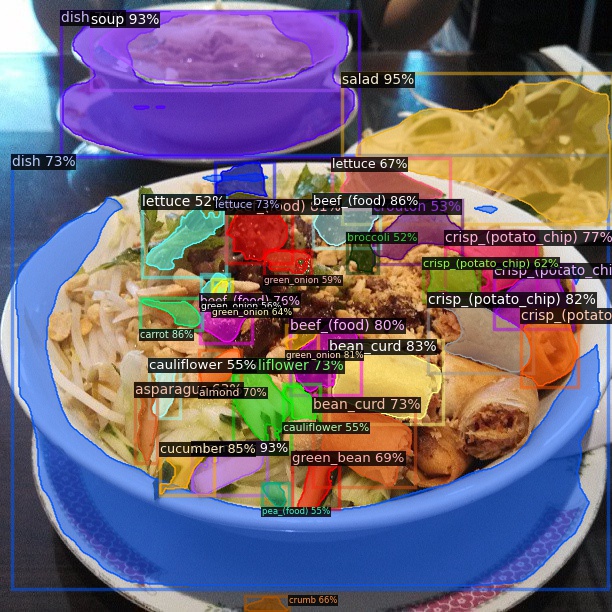} \hfill
    \mfigure{0.24}{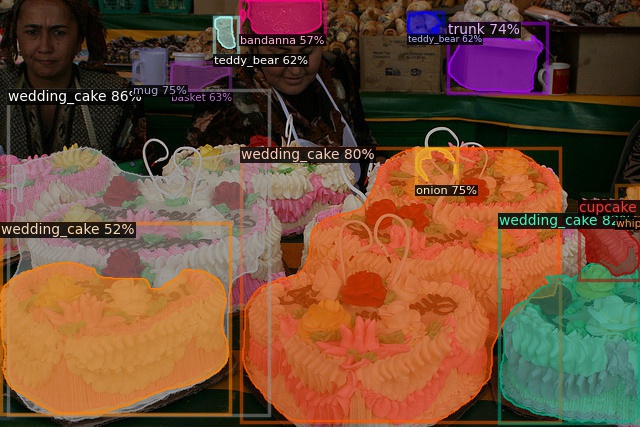} \hfill
    \mfigure{0.24}{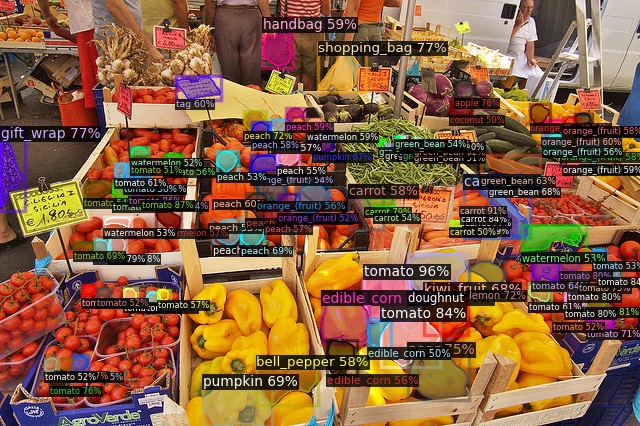}\vspace{-2mm}\\
$\underbracket[1pt][2.0mm]{\hspace{\textwidth}}_%
    {\substack{\vspace{-3.0mm}\\\colorbox{white}{~~Mask RCNN with DropLoss~~}}}$\vspace{1mm}\\%

  \caption{
  Visual results comparison. Qualitative results of the Mask R-CNN trained with standard cross-entropy loss (\emph{Top}) and the proposed DropLoss (\emph{Bottom}). Instances with scores larger than 0.5 are shown. 
  }
  \label{fig:visual}
\end{figure*}

\paragraph{Comparisons with state-of-the-art methods.}

In our experiments, we use Mask R-CNN \citep{he2017mask} as our architecture and compare it with two baseline training methods: standard Mask R-CNN and the equalization loss \citep{tan2020eql}. 
To verify that DropLoss is effective across different settings, we validate on several different architectures and backbones. 
We test ResNet50 and ResNet101 \citep{he2016resnet} as backbones, and compare the Cascades R-CNN \citep{cai2018cascade} as an alternative architecture to Mask R-CNN \cite{HeGDG20}. 
Table \ref{tab:LVIS Effectiveness} reports the results, where all methods are tested using the same experiment settings and environment. 
We find that DropLoss achieves improved performance (in terms of overall AP) compared with both baselines across all backbones and architectures.
We are most interested in the AP$_\mathrm{r}$, AP$_\mathrm{c}$, AP$_\mathrm{f}$ and AR. 
Although the AP$_\mathrm{f}$ (frequent) decreases slightly in our method, our AP$_\mathrm{r}$ (rare) and AP$_\mathrm{c} (common)$ increase significantly. 
Our method improves the AP and AR by a large margin, indicating the overall performance across all categories is improved. 
In particular, using Mask R-CNN with ResNet-50 as the backbone, we achieve a 1.7 AP improvement over the state-of-the-art method~\cite{tan2020eql} (winner of the LVIS 2019 challenge).
Across all the settings, compared with the baselines, DropLoss can more successfully balance the tradeoff between rare and frequent categories, resulting in better performance in the long-tailed distribution dataset.

\topic{Incorporating with Resampling Methods.}
Here we show that our approach can be combined with state-of-the-art resampling methods to improve learning long-tailed distribution further. 
Specifically, we adopt the Repeat Factor Sampling (RFS) \cite{GuptaDG19} that uses the number of images per category to determine the sampling frequency. 
\tabref{compare} shows the quantitative comparisons of different loss function choices on the LVIS v0.5 validation set.\footnote{Note that the EQL results (25.5 AP) are not consistent with the reported results (26.1 AP). We use the public implementation without changes and report the average over 3 runs. The difference may be due to number of GPUs used for training, resulting in 
different batch normalization. For fair comparisons, we 
use the same hardware and experimental setting to train all models.
}
We find that applying RFS generally improves the performance of all the methods. 
The proposed DropLoss compares favorably against other baseline methods either with or without using RFS.
Note that the RFS method rebalances based on \emph{overall} data distribution, while DropLoss reweights the loss based on statistics in the \emph{each batch}. 
The complementary nature of the two methods may explain why integrating RFS and the DropLoss leads to improved results.

\topic{Measuring the Frequent-rare Category Performance Tradeoff.}
%
Methods for learning long-tail distribution often involve a tradeoff between accuracy on rare, common, and frequent categories.
Here we wish to quantify this tradeoff for various methods. 
We compare our proposed DropLoss against three baselines: equalization loss \cite{tan2020eql}, background equalization loss, and fixed drop ratio.

Equalization loss and DropLoss have no tunable hyperparameters. 
Background equalization loss has the log base as a tunable hyperparameter. 
A fixed drop ratio has the drop ratio as a hyperparameter.
These methods may be adjusted to measure the tradeoff between object categories.
We can use the Pareto Frontier from multi-objective optimization to visualize this tradeoff, as seen in Figure \ref{fig:parteo}. 
We observe that for reweighting methods with tunable hyperparameters, improvement in rare AP$_\mathrm{r}$ or common AP$_\mathrm{c}$ generally leads to a rapid decrease in frequent AP$_\mathrm{f}$. 
Our proposed DropLoss does not have tunable hyperparameters, but Figure \ref{fig:parteo} demonstrates that DropLoss balances more effectively between AP$_\mathrm{r}$, AP$_\mathrm{c}$ and AP$_\mathrm{f}$, resulting in higher overall AP than other baselines. 

DropLoss adapts to the sampling distribution so that if a rare category appears in a given batch, its loss is less likely to be dropped. 
However, if a rare category does not appear in a batch, the chance of its loss being dropped is very high. 
This allows the network to \emph{dynamically} attend to the categories that it sees in a given batch, decreasing drop loss probability selectively for only those categories. 
We postulate that this allows the network to achieve a better overall balance between frequent and infrequent categories.

\topic{Quantitative Comparison Between the DropLoss and Our Proposed Baseline BEQL.} To validate that the DropLoss provides better pareto-efficiency over BEQL, in \tabref{compare_beql}, we compare the DropLoss with two \emph{best-performing} results (in term of overall AP) from BEQL. 
DropLoss still offers overall better performance (not only in AP but also in AR) despite not sweeping the parameters on the validation set to find the best performance as in BEQL.

\begin{table}[htbp]
\scriptsize
\centering
\resizebox{\linewidth}{!}{%
\begin{tabular}{@{}l@{\,}c@{\;}c@{\;\,}c@{\;\,}c@{\;\,}c@{\;\,}c@{\;\,}c@{\;}c@{\;}c@{}}
\toprule
 Method & AP (\%) & AP$_\mathrm{r}$ & AP$_\mathrm{c}$ & AP$_\mathrm{f}$ & AP$_\mathrm{bbox}$ & AR \\ \midrule
BEQL ($b=4$) & 25.2 & \textbf{14.6} & \textbf{28.1} & 25.9  & 24.9 & 34.1 \\
BEQL ($b=5$) & 25.1 & 13.4 & 27.4 & 26.9 & 24.8 & 34.3 \\
DropLoss & \textbf{25.5} & 13.2 & 27.9 & \textbf{27.3} & \textbf{25.1} & \textbf{34.8} \\ \bottomrule
\end{tabular}%
}
\caption{Comparison between DropLoss and two top-performing results from our another proposed method BEQL. Evaluation on LVIS v0.5 validation set. DropLoss offers overall better performance (in both AP and AR) compared with the BEQL baseline.
\label{tab:compare_beql}}
\end{table}

\topic{Visual Results.}
\figref{visual} demonstrates the results on a dense instance segmentation example containing common/rare category.
For example, the goose in the first image is a `common' object category. 
We demonstrate the suppression of these less-frequent categories, as most of the geese in this image are classified as background or with low confidence. 
In contrast, training the model with the proposed loss correctly identifies all geese as foreground, and predicts category ``goose'' with high confidence and other waterbirds with lower confidence.
Despite the stochastic removal of rare and common category losses for background proposals, we find that the network does not misclassify background regions as foreground. 
The distinction between background and foreground is likely less difficult to learn than the distinction between foreground image categories, so reducing background gradients does not appear to significantly affect background/foreground classification. 
By reducing the suppression of rare and common categories via background predictions, our method allows for rare and common categories to improve prediction scores, decreasing bias towards frequent categories.



\section{Conclusions}

Through analysis of the loss gradient distributions over rare, common, and frequent categories, we discovered that disproportionate background gradients suppress less-frequent categories in the long-tailed distribution problem.
To address this problem, we propose DropLoss, which balances the background loss gradients between different categories via random sampling and reweighting. 
Our method provides a sizable performance improvement across different backbones and architectures by improving the balance between object categories in the long-tail instance segmentation setting.
While we focus on the challenging problem of instance segmentation in our experiments, we expect that DropLoss may be applicable to other visual recognition problems with long-tailed distributions. 
We leave the exploration to other tasks as future work.

\section{Acknowledgments}

This work was supported in part by Virginia Tech, the Ministry of Education BioPro A+, and MOST Taiwan under Grant 109-2634-F-001-012.
We are particularly grateful to the National Center for High-performance Computing for computing time and facilities.

\bibliography{references} 
\end{document}